\documentclass[sigconf]{acmart}
\AtBeginDocument{%
  }

\copyrightyear{2026}
\acmYear{2026}
\setcopyright{cc}
\setcctype{by-nc-nd}
\acmConference[MM '26]{Proceedings of the 34th ACM International Conference on Multimedia}{November 10--14, 2026}{Rio de Janeiro, Brazil}
\acmBooktitle{Proceedings of the 34th ACM International Conference on Multimedia (MM '26), November 10--14, 2026, Rio de Janeiro, Brazil}
\acmDOI{10.1145/3767308.3835975}
\acmISBN{979-8-4007-2213-4/2026/11}

\begin{document}

\title{DMM-Align: Closed-Loop Optimization for 2D-3D Registration with Dual-Role Diffusion}


\author{Chongjian Wang}
\orcid{0009-0008-7204-3225}

\affiliation{%
  \institution{Shandong Women's University}
  \city{Jinan}
  \country{China}
}

\affiliation{%
  \institution{Shandong University of Science and Technology}
  \city{Qingdao}
  \country{China}
}

\email{202311080223@sdust.edu.cn}

\author{Junjie Gao}
\authornote{Corresponding author.}
\orcid{0000-0001-7087-0886}

\affiliation{%
  \institution{Shandong Women's University}
  \city{Jinan}
  \country{China}
}

\email{junjie.gao@sdwu.edu.cn}

\renewcommand{\shortauthors}{Chongjian Wang and Junjie Gao}


\begin{abstract}
    2D–3D registration remains brittle in challenging scenarios such as low overlap, occlusion, repetitive structures, and severe cross-modal ambiguity. A key reason is that existing methods improve representation learning, correspondence estimation, or pose computation in isolation, while the dominant failure mode is inherently cross-level, where errors propagate between features, correspondences, and pose. To address this limitation, we propose \textbf{DMM-Align:} \textbf{D}iffusion-based \textbf{M}atching \textbf{M}atrix \textbf{Align}ment, a closed-loop framework that couples correspondence refinement, pose estimation, and representation learning through a shared differentiable geometric state. Our method leverages diffusion in two coordinated roles: a geometry-aware diffusion process refines the soft matching matrix for robust correspondence estimation, while a geometry-conditioned diffusion teacher injects pose-induced supervision back into feature learning. These processes are connected via a differentiable geometric hinge that converts correspondences into a global pose and exposes geometric inconsistency to upstream modules. Extensive experiments on 7-Scenes and RGB-D Scenes V2 demonstrate that DMM-Align consistently outperforms strong baselines, especially under low-overlap and heavy-occlusion conditions, highlighting the effectiveness of closed-loop geometric feedback for robust 2D–3D registration.
\end{abstract}




\maketitle

\section{Introduction}

Establishing reliable correspondences between images and point clouds is a fundamental problem in computer vision, with broad applications in visual localization~\cite{Selvaraju2016GradCAMVE,Zhou2024TheNM}, augmented reality~\cite{Azuma1997ASO}, robotics~\cite{Huang2022VisualLM,Patle2019ARO}, SLAM~\cite{DurrantWhyte2006SimultaneousLA}, and 3D reconstruction~\cite{Hong2023LRMLR,Leroy2024GroundingIM}. Given an image and a 3D scene representation, 2D--3D registration estimates the camera pose by identifying geometrically consistent pixel--point correspondences and solving for their relative transformation. Correspondence quality is therefore crucial: inaccurate matches directly degrade pose recovery, whereas robust correspondences enable accurate and stable registration under substantial viewpoint and modality differences.

\begin{figure}[t]
\centering
\includegraphics[width=\linewidth]{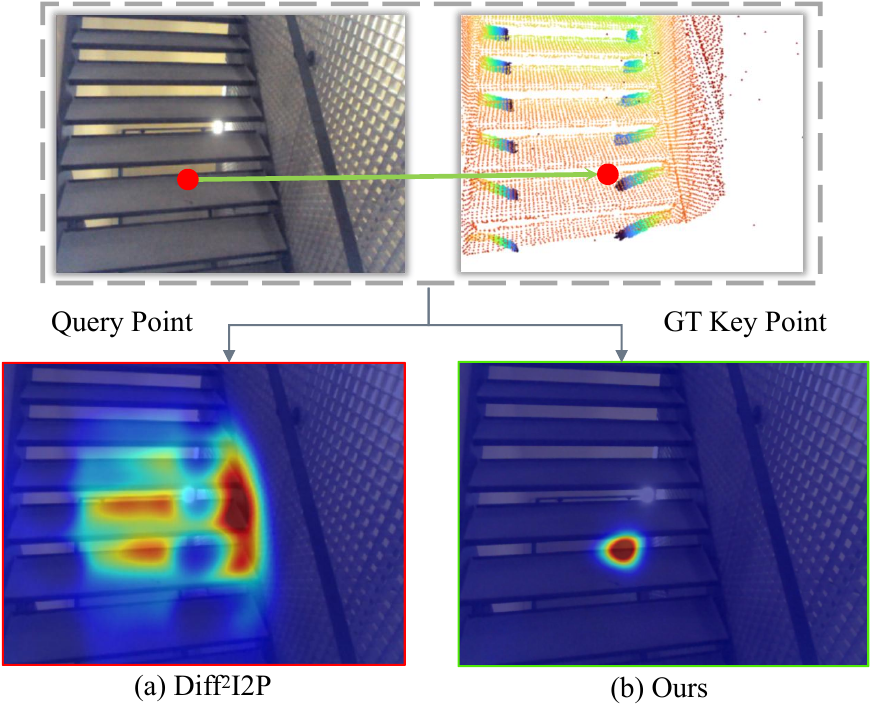}
\vspace{-20pt}
\caption{Comparison of correspondence confidence maps for a query 3D point. We select a point in a structurally repetitive region and visualize its matching confidence over the image as predicted by Diff2I2P and our method. Our method produces a response that is more concentrated near the ground-truth correspondence.}
\label{fig:1}
\vspace{-10pt}
\end{figure}

Despite substantial progress, 2D--3D registration remains brittle under low overlap, heavy occlusion, repetitive structures, and severe cross-modal ambiguity. Existing methods address these challenges from different perspectives. Some approaches~\cite{Feng20192D3DMatchnetLT,Pham2019LCDLC,Wang2021P2NetJD} learn stronger cross-modal representations to bridge 2D appearance and 3D geometry. Others focus on correspondence reasoning through coarse-to-fine matching~\cite{Li20232D3DMATR2M}, structural refinement~\cite{mu2025diff2i2p}, or iterative optimization of soft correspondence matrices~\cite{Wu2024DiffRegDM}. More recently, diffusion-based techniques~\cite{wang2023freereg,mu2025diff2i2p} have been introduced to improve matching quality or provide stronger priors for feature learning. Although these advances significantly improve robustness, most methods still treat representation learning, correspondence estimation, and pose computation as loosely coupled stages in a largely feed-forward pipeline.

This design leaves an important limitation insufficiently addressed. In challenging cases, registration errors rarely originate from a single stage. Instead, they arise through \emph{cross-level error propagation}: weak features bias correspondence estimation, corrupted correspondences destabilize pose recovery, and inaccurate poses provide unreliable geometric signals for improving upstream representations. Local errors can therefore accumulate across stages, while the system lacks an effective mechanism to feed global geometric evidence back into earlier components. Consequently, improving an individual module does not necessarily result in globally consistent registration. As illustrated in Fig.~\ref{fig:1}, existing methods~\cite{mu2025diff2i2p} may produce diffuse responses in repetitive or ambiguous regions, whereas accurate registration requires confidence to be concentrated near the true correspondence. This mismatch between local correspondence prediction and global geometric consistency is the problem addressed in this work.

We argue that a central bottleneck of 2D--3D registration is the absence of a \emph{closed-loop optimization mechanism} that explicitly couples representation learning, correspondence refinement, and pose estimation. To this end, we propose \textbf{DMM-Align}, a unified framework that connects these stages through a shared \emph{differentiable geometric state}. Its key idea is that refined correspondences should not only improve pose estimation, but the resulting pose-induced geometric consistency should also flow back to supervise feature learning. Local correspondence corrections are thus lifted into a global pose estimate, while global geometric evidence reshapes the representations used to generate subsequent correspondences.

DMM-Align integrates three tightly connected components. First, we introduce \emph{geometry-aware matching matrix diffusion}, which refines soft 2D--3D correspondences through iterative denoising under geometric guidance, producing more reliable matching structures under ambiguity and noise. Second, we develop a \emph{differentiable geometric hinge} that converts refined correspondences into a trainable global pose state, allowing pose-level geometric inconsistency to be propagated to upstream modules. Third, we design \emph{geometry-conditioned diffusion distillation}, which feeds pose-induced geometric cues back into feature learning and provides dense, structured, and uncertainty-aware supervision. Diffusion therefore plays two coordinated but distinct roles: it acts as both a correspondence optimizer and a representation teacher. Rather than functioning as isolated components, the two diffusion processes are coupled through the shared geometric state to form a unified closed-loop optimization framework.

An important advantage of DMM-Align is that it improves registration not by simply stacking stronger modules, but by enabling interaction across different levels of the pipeline. Correspondence refinement, pose estimation, and representation learning can benefit from one another during training, leading to more stable optimization and stronger geometric consistency. Extensive experiments on 7-Scenes~\cite{Glocker2013RealtimeRC} and RGB-D Scenes V2~\cite{Lai2014UnsupervisedFL} show that DMM-Align consistently outperforms strong baselines, with particularly clear improvements under low-overlap and heavy-occlusion conditions. Comprehensive ablations further demonstrate that these gains arise from the proposed feedback coupling rather than merely from combining multiple powerful components.

Our contributions are summarized as follows:
\begin{itemize}
\item We identify the lack of cross-level geometric feedback as a key limitation of existing 2D--3D registration methods and reformulate registration as a closed-loop optimization process across representation learning, correspondence estimation, and pose recovery.
\item We propose \textbf{DMM-Align}, a unified framework that connects these stages through a shared differentiable geometric state, allowing local correspondence corrections and global pose supervision to interact within the same optimization loop.
\item We introduce a dual-role diffusion design, comprising geometry-aware matching matrix diffusion for correspondence refinement and geometry-conditioned diffusion distillation for pose-guided representation learning.
\item Extensive experiments and ablations demonstrate that DMM-Align achieves superior robustness and accuracy on challenging benchmarks, especially under low-overlap and heavy-occlusion settings.
\end{itemize}

\section{Related Work}

\subsection{Image and Point Cloud Registration}

Image and point cloud registration provide the technical foundation for 2D--3D alignment. Classical image registration typically follows a detect-then-match pipeline~\cite{Lowe1999ObjectRF,Rublee2011ORBAE,DeTone2017SuperPointSI,Dusmanu2019D2NetAT,Luo2020ASLFeatLL,Sarlin2019SuperGlueLF}, where sparse keypoints are detected and described using handcrafted or learned features, followed by geometric verification and pose recovery. More recent methods~\cite{Lee2021PatchMatchBasedNC,Li2020CorrespondenceNW,Rocco2018NeighbourhoodCN,Rocco2020EfficientNC,Sun2021LoFTRDL,Zhou2020Patch2PixEP} adopt detector-free formulations that infer correspondences directly from dense features, often with coarse-to-fine refinement~\cite{Sun2021LoFTRDL}.

Point cloud registration has evolved from handcrafted descriptors such as PPF~\cite{Drost2010ModelGM} and FPFH~\cite{Rusu2009FastPF} to learning-based features~\cite{Yew20183DFeatNetWS,Deng2018PPFNetGC,Gojcic2018ThePM,Ao2020SpinNetLA}, detector-free matching~\cite{Choy2019FullyCG,Thomas2019KPConvFA}, hierarchical correspondence search~\cite{Yu2021CoFiNetRC}, and transformer-based feature interaction~\cite{Gao2023OAAFormerRA,Qin2023GeoTransformerFA,Yu2023RotationInvariantTF,Chen2024DynamicCT,Yao2024PARENetPR}.

However, these methods are generally developed under settings where both inputs share similar appearance or geometric structure. In cross-modal registration, the gap between image appearance and 3D geometry makes correspondence estimation more ambiguous, and strong local matching does not necessarily yield globally consistent poses. Moreover, geometric inconsistency from pose recovery is rarely reused to improve upstream representations. This motivates approaches that explicitly incorporate geometric feedback into correspondence learning.

\subsection{Cross-Modal Registration}

Compared to single-modality matching, establishing correspondences between images and point clouds is considerably more challenging due to the inherent gap between 2D appearance and 3D geometry.  Early approaches~\cite{Feng20192D3DMatchnetLT,Pham2019LCDLC,Wang2021P2NetJD} typically follow a detect-then-match paradigm, where keypoints are independently detected in each modality and matched via cross-modal descriptors.  However, defining repeatable and consistent keypoints across heterogeneous domains is inherently difficult, often leading to low inlier ratios and unstable performance.  To address these limitations, recent methods~\cite{Li2021DeepI2PIC,Li20232D3DMATR2M,Ren2022CorrI2PDI,Zhou2023DifferentiableRO,mu2025diff2i2p} increasingly adopt detector-free pipelines that establish correspondences through direct cross-modal feature interaction. 
Many of these methods further incorporate coarse-to-fine strategies to progressively refine correspondences~\cite{Li20232D3DMATR2M,mu2025diff2i2p}, where hierarchical matching and contextual feature aggregation first identify globally consistent regions and then refine them into accurate pixel--point correspondences.

Beyond direct matching, some approaches~\cite{wang2023freereg,mu2025diff2i2p,Wu2024DiffRegDM,Zhou2023DifferentiableRO} further introduce diffusion-based modeling, correspondence refinement, or differentiable geometric solvers to improve matching reliability and pose accuracy.  Despite these advances, most existing methods still focus primarily on improving matching quality within a feed-forward pipeline, where representation learning, correspondence estimation, and pose recovery are only loosely coupled.  As a result, global geometric cues are rarely reused as structured feedback for upstream modules, making cross-level error propagation insufficiently addressed under challenging conditions such as low overlap, occlusion, and repetitive structures.

\subsection{Diffusion for Registration and Matching}

Diffusion models~\cite{ho2020denoising,Song2020ScoreBasedGM} are well suited to geometric prediction under uncertainty because they iteratively refine noisy structured variables. They have been applied to SE(3) pose generation~\cite{jiang2023se}, 6D pose estimation~\cite{Xu20236DDiffAK}, overlap refinement, and point cloud registration~\cite{She2024PointDifformerRP}, where gradual denoising provides a robust alternative to one-shot prediction.

Diffusion has also been introduced into cross-modal registration to model uncertain correspondences or provide stronger feature priors~\cite{mu2025diff2i2p}. Existing methods~\cite{mu2025diff2i2p,Wu2024DiffRegDM}, however, generally use diffusion for either correspondence refinement or representation enhancement, improving only one stage at a time. In contrast, DMM-Align assigns diffusion two coordinated roles: refining the soft matching matrix under geometric guidance and supervising representation learning with pose-conditioned signals. Coupling both processes through a shared differentiable geometric state explicitly links local correspondence correction with global pose feedback.

\section{Method}

\begin{figure*}[t]
\vspace{-4pt}
  \centering
  \includegraphics[width=\textwidth]{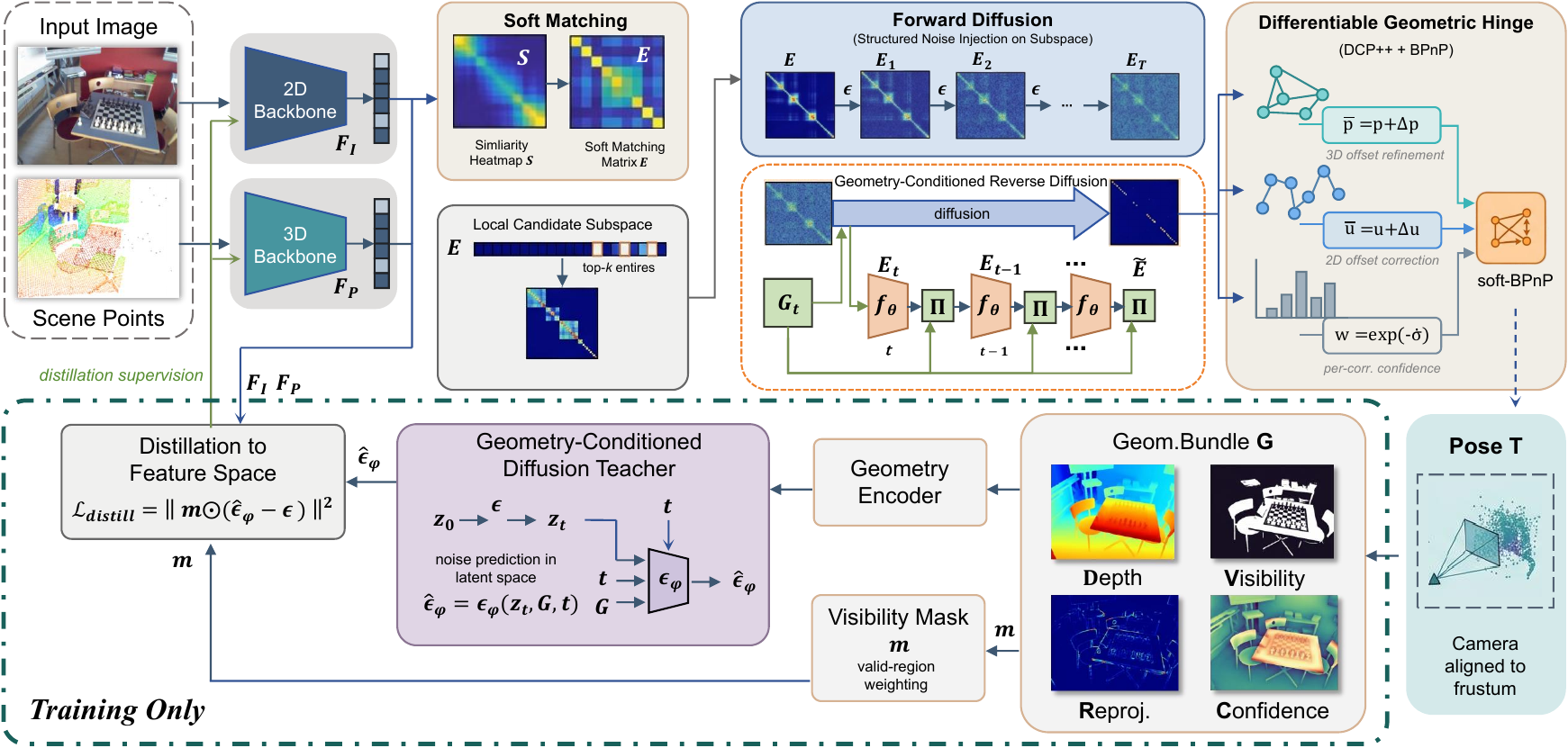}
    \caption{\textbf{Overview of DMM-Align.}
    Given a query image and a scene point cloud, a detection-free coarse-to-fine backbone extracts multi-scale features and constructs an initial soft 2D--3D matching matrix.
    A geometry-aware diffusion module refines the correspondence structure within local candidate subspaces.
    The refined matches are converted into a global pose via a differentiable geometric hinge, which also produces pose-induced cues such as depth, visibility, and reprojection consistency.
    These cues are fed into a geometry-conditioned diffusion teacher to distill pose-level supervision back to the feature space, forming a closed-loop optimization across features, correspondences, and pose.
    During inference, the distillation branch is removed, and the pipeline reduces to feature extraction, matching refinement, and pose estimation.}

  \label{fig:2}
\vspace{-8pt}
\end{figure*}

\subsection{Overview}

Given a query image $I \in \mathbb{R}^{H \times W \times 3}$, a scene point cloud
$P=\{x_i\}_{i=1}^{N}$ with $x_i \in \mathbb{R}^3$, and camera intrinsics $K$, the goal of 2D--3D registration is to estimate the camera pose $T=[R|t]\in SE(3)$, where $R\in SO(3)$ and $t\in\mathbb{R}^3$. A standard correspondence-based pipeline first predicts 2D--3D matches
\[
\mathcal{C}=\{(x_k,u_k)\}_{k=1}^{M},
\]
and then estimates the pose by minimizing the reprojection error
\begin{equation}
\min_{R,t}
\sum_{(x_k,u_k)\in\mathcal{C}}
\left\|
\pi\!\left(K(Rx_k+t)\right)-u_k
\right\|_2^2,
\end{equation}
where $\pi(\cdot)$ denotes projection from 3D space to the image plane. Although solvable via PnP-RANSAC~\cite{Fischler1981RandomSC,lepetit2009epnp}, the performance is highly sensitive to correspondence quality: unreliable matches quickly degrade pose estimation, while the resulting pose provides limited supervision to upstream modules.

We observe that failures in 2D--3D registration are inherently \emph{cross-level}: ambiguous features lead to diffuse correspondences, unstable correspondences distort pose estimation, and pose errors hinder meaningful geometric feedback. Existing methods typically improve these components in isolation, resulting in a largely feed-forward pipeline. 

We instead formulate registration as a \emph{closed-loop optimization process} over four coupled variables:
\[
\mathbf{F}\;\rightarrow\;\mathbf{E}\;\rightarrow\;\mathbf{T}\;\rightarrow\;\mathbf{G}\;\rightarrow\;\mathbf{F},
\]
where $\mathbf{F}=\{\mathbf{F}^I,\mathbf{F}^P\}$ denotes image and point representations, $\mathbf{E}$ is a soft matching matrix, $\mathbf{T}$ is the estimated pose, and $\mathbf{G}$ denotes pose-induced geometry. As shown in Fig.~\ref{fig:2}, DMM-Align instantiates this loop with three components: geometry-aware diffusion for correspondence refinement, a differentiable geometric hinge for pose estimation, and a geometry-conditioned diffusion teacher for representation learning, enabling joint optimization of correspondences, pose, and features within a unified training framework.

\subsection{Feature Extraction and Soft Matching}

We build DMM-Align on top of a detection-free coarse-to-fine 2D--3D matcher. An image encoder $E_I$ extracts multi-scale visual features from the query image, while a point-cloud encoder $E_P$ extracts hierarchical geometric features from the scene point cloud. In practice, we use a ResNet-FPN\cite{He2015DeepRL} backbone for the image branch and a KPConv-FPN\cite{Thomas2019KPConvFA} backbone for the point branch.

Let $\hat{\mathbf{F}}^I \in \mathbb{R}^{M_c \times d}$ and $\hat{\mathbf{F}}^P \in \mathbb{R}^{N_c \times d}$ denote the coarse-level image and point descriptors. We first compute a coarse similarity matrix
\begin{equation}
\mathbf{S}^c = \hat{\mathbf{F}}^I (\hat{\mathbf{F}}^P)^\top,
\end{equation}
whose entries measure the affinity between image patch tokens and point-cloud patch tokens. Based on $\mathbf{S}^c$, we select candidate coarse pairs and restrict subsequent fine matching to these local subspaces, reducing ambiguity and computational cost.

At the fine level, let $\mathbf{F}^I \in \mathbb{R}^{M \times d}$ and $\mathbf{F}^P \in \mathbb{R}^{N \times d}$ denote the image and point features used for local matching. Their pairwise affinities define
\begin{equation}
\mathbf{S} = \mathbf{F}^I (\mathbf{F}^P)^\top.
\end{equation}
Instead of collapsing these affinities into hard correspondences, we normalize them into a soft matching matrix
\begin{equation}
\mathbf{E} = \operatorname{Sinkhorn}(\mathbf{S}),
\end{equation}
which approximates a partial doubly stochastic correspondence matrix. To account for unmatched elements caused by occlusion, truncation, or incomplete overlap, we further augment the assignment with dustbin entries.

This soft formulation preserves uncertainty that would be lost under hard top-$1$ matching and exposes correspondence as a differentiable structured variable. As a result, $\mathbf{E}$ serves as the central latent state connecting feature learning and pose estimation, and can be further refined by geometric reasoning in subsequent stages.

\subsection{Geometry-Aware Matching Matrix Diffusion}

Predicting $\mathbf{E}$ solely from feature similarity is often unreliable in challenging scenes. Repetitive structures, textureless regions, occlusion, and low overlap can all produce diffuse or inconsistent confidence patterns. To address this issue, we treat the soft correspondence matrix as a structured variable and refine it through a geometry-aware diffusion process.

Starting from the initial matching matrix $\mathbf{E}$, we define the forward process as
\begin{equation}
\mathbf{E}_t
=
\sqrt{\bar{\alpha}_t}\,\mathbf{E}
+
\sqrt{1-\bar{\alpha}_t}\,\boldsymbol{\epsilon},
\qquad
\boldsymbol{\epsilon}\sim\mathcal{N}(0,\mathbf{I}),
\end{equation}
where $t$ is the diffusion timestep and $\bar{\alpha}_t$ is the cumulative noise schedule. Since $\mathbf{E}$ lies in a constrained assignment space, direct noising may destroy its structure. We therefore apply row-wise and column-wise stabilization after each step to preserve numerical validity and relative correspondence patterns.

The reverse process is implemented by a denoising network $f_\theta$:
\begin{equation}
\hat{\boldsymbol{\epsilon}}
=
f_\theta(\mathbf{E}_t,\mathbf{G}_t,t),
\end{equation}
where $\mathbf{G}_t$ denotes geometry-aware conditioning signals derived from the current registration state, including reprojection inconsistency, depth ordering, visibility, and neighborhood support. Conditioning on $\mathbf{G}_t$ allows the denoiser to prefer correspondence structures that are globally consistent with the current pose rather than merely sharpening local appearance affinity.

For efficiency, diffusion is performed only on local candidate submatrices constructed from the top-$k$ point candidates for each image token within the coarse-to-fine matching region. This is sufficient in practice, since ambiguity is typically concentrated among a small set of competing hypotheses.

The diffusion module is trained with the standard denoising loss
\begin{equation}
\mathcal{L}_{\text{diff}}
=
\mathbb{E}_{t,\boldsymbol{\epsilon}}
\left[
\left\|
\hat{\boldsymbol{\epsilon}}-\boldsymbol{\epsilon}
\right\|_2^2
\right].
\end{equation}
The resulting refined matching matrix $\tilde{\mathbf{E}}$ is then passed to the pose estimation stage.

\subsection{Differentiable Geometric Hinge}

After refining the soft correspondence matrix, we convert it into a global pose through a differentiable geometric hinge. Unlike a hard solver such as PnP-RANSAC, which breaks gradient flow, this module transforms local soft correspondences into a trainable pose state while exposing geometric inconsistency to upstream modules.

Given the refined matrix $\tilde{\mathbf{E}}$, each non-negligible entry defines a weighted 2D--3D match:
\[
\mathcal{C}_w
=
\{(x_j,u_i,w_{ij})\},
\qquad
w_{ij}=\tilde{\mathbf{E}}_{ij},
\]
where $u_i$ is the image location of the $i$-th token and $x_j$ is the coordinate of the $j$-th 3D point. Since local feature matching may still be slightly biased, we further predict lightweight residual corrections for both modalities:
\begin{equation}
\Delta x_{ij}
=
\operatorname{MLP}_{x}\!\left(
[f_i^I,f_j^P,x_j,K^{-1}(u_i)]
\right),
\end{equation}
\begin{equation}
\Delta u_{ij}
=
\operatorname{MLP}_{u}\!\left(
[f_i^I,f_j^P,x_j,K^{-1}(u_i)]
\right),
\end{equation}
where $f_i^I$ and $f_j^P$ are the corresponding fine-level features. These residuals correct small geometric deviations before pose estimation, while the soft weights retain uncertainty from the refined matching matrix.

The corrected correspondences are then passed to a differentiable weighted PnP solver:
\begin{equation}
\mathbf{T}
=
\operatorname{DPnP}
\big(
\{(x_j+\Delta x_{ij},\,u_i+\Delta u_{ij},\,w_{ij})\}
\big).
\end{equation}
In practice, we adopt a BPnP-style solver for stable gradient propagation. To prevent the model from relying excessively on residual offsets, we regularize them by
\begin{equation}
\mathcal{L}_{\text{off}}
=
\sum_{i,j}
\left(
\|\Delta x_{ij}\|_2^2
+
\|\Delta u_{ij}\|_2^2
\right),
\end{equation}
and supervise the estimated pose with
\begin{equation}
\mathcal{L}_{\text{pose}}
=
\|R-R^{\text{gt}}\|_F^2
+
\|t-t^{\text{gt}}\|_2^2.
\end{equation}

Besides pose estimation, the hinge also produces the geometric state required for closing the loop. Once a pose is obtained, we compute pose-induced signals such as depth, visibility, and reprojection error, which form $\mathbf{G}$ and are later used for both correspondence refinement and representation supervision.

\subsection{Geometry-Conditioned Distillation}

A key difference between DMM-Align and conventional feed-forward matchers is that the estimated pose is not treated as the endpoint of the pipeline. Instead, we convert the current pose into structured geometric cues and feed them back to the representation space through a geometry-conditioned diffusion teacher.

Given the estimated pose $\mathbf{T}=[R|t]$, we transform the scene point cloud into the camera frame and project it to the image plane, yielding a pose-induced geometric bundle
\[
\mathbf{G}=\{D,V,R,C\},
\]
where $D$ denotes the depth map, $V$ the visibility mask, $R$ the reprojection residual, and $C$ a geometry-aware confidence map. Since direct projection produces sparse depth, we introduce a differentiable densification operator $\mathcal{F}$:
\begin{equation}
D
=
\mathcal{F}\big(
\pi(K(RP+t))
\big).
\end{equation}

We then employ a geometry-conditioned diffusion teacher to distill pose-level information into the representation space. Let $z$ be the latent encoding of the input image and $z_t$ a noisy latent at timestep $t$. The teacher predicts the injected noise conditioned on geometry:
\begin{equation}
\hat{\epsilon}_{\phi}
=
\epsilon_{\phi}(z_t,\mathbf{G},t),
\end{equation}
where $\phi$ denotes the teacher parameters. We adopt a score-distillation style objective:
\begin{equation}
\mathcal{L}_{\text{distill}}
=
\mathbb{E}_{t,\epsilon}
\left[
\|m\odot(\hat{\epsilon}_{\phi}-\epsilon)\|_2^2
\right],
\end{equation}
where $m$ is a visibility mask to suppress supervision in invalid regions.

This module converts global pose quality into local representation supervision. Geometrically consistent poses reinforce reliable features, while uncertain regions receive softer guidance through diffusion and masking, enabling stable cross-level optimization.

\subsection{Loss Functions}

The overall objective of DMM-Align combines correspondence refinement, pose estimation, residual regularization, and representation feedback:
\begin{equation}
\mathcal{L}
=
\lambda_1 \mathcal{L}_{\text{diff}}
+
\lambda_2 \mathcal{L}_{\text{pose}}
+
\lambda_3 \mathcal{L}_{\text{off}}
+
\lambda_4 \mathcal{L}_{\text{distill}},
\end{equation}
where $\lambda_1,\lambda_2,\lambda_3,\lambda_4$ balance the contributions of each term.

Specifically, $\mathcal{L}_{\text{diff}}$ supervises the denoising of the matching matrix, $\mathcal{L}_{\text{pose}}$ enforces global geometric accuracy, $\mathcal{L}_{\text{off}}$ regularizes residual corrections in the geometric hinge, and $\mathcal{L}_{\text{distill}}$ propagates pose-induced supervision back to feature learning.

We train the entire framework end-to-end. During inference, the geometry-conditioned diffusion teacher is removed, while feature extraction, matching refinement, and pose estimation remain unchanged, preserving the efficiency of the original pipeline.

\section{Experiments}

We evaluate DMM-Align from four complementary perspectives.
First, we compare it with representative prior methods on standard 2D--3D registration benchmarks.
Second, we test whether the gain truly arises from the proposed closed-loop feedback mechanism rather than from the coexistence of several individually strong components.
Third, we conduct ablation studies to examine whether the key structural designs are necessary for making the loop effective.
Finally, we analyze efficiency and controllability to assess the practical value of the resulting system.

Overall, the experiments are designed not only to report final performance, but also to test the central hypothesis of this work:
coupling correspondence refinement, pose estimation, and representation learning through a differentiable geometric feedback pathway leads to stronger and more robust 2D--3D registration.

\subsection{Implementation Details}
\noindent\textbf{Network architecture.}
We build DMM-Align on top of a detection-free coarse-to-fine 2D--3D matching pipeline. Following strong prior methods such as 2D3D-MATR~\cite{Li20232D3DMATR2M} and Diff2I2P~\cite{mu2025diff2i2p}, we use a 4-stage ResNet-FPN~\cite{He2015DeepRL} as the image encoder and a 4-stage KPConv-FPN~\cite{Thomas2019KPConvFA} as the point cloud encoder to extract multi-scale features. An initial soft 2D--3D matching matrix is constructed within coarse-to-fine candidate regions, then refined by geometry-aware matching diffusion over local top-$k$ candidate subspaces. The refined correspondences are converted into a trainable pose estimate through the differentiable geometric hinge. During training, pose-induced geometric cues are further injected into the feature space through geometry-conditioned distillation, while at inference time this distillation branch is removed and the pipeline reduces to feature extraction, matching refinement, and pose estimation.

\noindent\textbf{Datasets.}
We evaluate DMM-Align on two public indoor benchmarks: 7-Scenes~\cite{Glocker2013RealtimeRC} and RGB-D Scenes V2~\cite{Lai2014UnsupervisedFL}. These two datasets stress complementary aspects of the problem. 7-Scenes is a standard benchmark for indoor visual relocalization and allows direct comparison with prior 2D--3D registration methods. RGB-D Scenes V2 is structurally more diverse, with larger viewpoint variation, heavier occlusion, and stronger cross-modal ambiguity, making it particularly suitable for testing whether the proposed closed-loop mechanism generalizes beyond a single benchmark style.

\noindent\textbf{Metrics.}
On 7-Scenes, we report Inlier Ratio (IR), Feature Matching Recall (FMR), and Registration Recall (RR). On RGB-D Scenes V2, we report Patch Inlier Ratio (PIR), IR, FMR, and RR. These metrics jointly evaluate correspondence quality, matching robustness, and final registration performance. For efficiency analysis, we additionally report inference time, model size, and GPU memory usage.

\noindent\textbf{Baselines.}
We compare DMM-Align with representative 2D--3D registration baselines, including FCGF-2D3D~\cite{Choy2019FullyCG}, Predator-2D3D~\cite{Huang2020PREDATORRO}, P2-Net~\cite{Wang2021P2NetJD}, 2D3D-MATR~\cite{Li20232D3DMATR2M}, Diff$^2$I2P~\cite{mu2025diff2i2p}, and FreeReg~\cite{wang2023freereg}.
These methods represent key lines of prior work, including feature matching, detector-free registration, and diffusion-based approaches.
Unless otherwise specified, all ablation studies are conducted on 7-Scenes.

\subsection{Evaluations on 7-Scenes}

\noindent\textbf{Quantitative results.}
We first evaluate DMM-Align on 7-Scenes\cite{Glocker2013RealtimeRC}, a standard benchmark for testing generalization across viewpoints in known scenes. The quantitative results are reported in Tab.~\ref{tab:1}. Overall, DMM-Align consistently outperforms prior methods across all three groups of metrics. For Inlier Ratio, DMM-Align achieves the best mean performance of 54.5\%, surpassing both MATR\cite{Li20232D3DMATR2M} and Diff$^2$I2P\cite{mu2025diff2i2p}. This indicates that the proposed framework produces more accurate and reliable pixel--point correspondences. For Feature Matching Recall, our method also achieves the best mean value of 92.5\%, showing that the improved correspondence quality generalizes across image--point-cloud pairs rather than only affecting a small number of examples. Most importantly, for Registration Recall, DMM-Align reaches 86.2\%, outperforming Diff$^2$I2P\cite{mu2025diff2i2p} by 3.2\% and MATR\cite{Li20232D3DMATR2M} by 10.4\%. The gain is not concentrated on only a few easy scenes. Instead, it is particularly clear on more difficult cases such as \emph{Heads} and \emph{Stairs}, where ambiguity, repeated structures, and unstable local evidence make stage-wise registration particularly brittle. For example, compared with Diff$^2$I2P\cite{mu2025diff2i2p}, DMM-Align improves RR from 74.0\% to 80.6\% on \emph{Heads} and from 36.5\% to 45.9\% on \emph{Stairs}. These results suggest that explicitly coupling correspondence refinement, pose estimation, and representation learning leads to a more reliable and globally consistent registration process.

\begin{table}[h]
\caption{Registration results of DMM-Align and baselines on 7-Scenes. The best result for each metric is shown in \textbf{bold}.}
\label{tab:1}
\centering
\footnotesize
\setlength{\tabcolsep}{3.6pt}
\renewcommand{\arraystretch}{1.1}
\begin{tabular}{lcccccccc}
\toprule
Model & Chess & Fire & Heads & Office & Pump & Kitchen & Stairs & Mean \\
\midrule
\multicolumn{9}{c}{Inlier Ratio $\uparrow$} \\
\midrule
FCGF\cite{Choy2019FullyCG}                & 34.2 & 32.8 & 14.8 & 26.0 & 23.3 & 22.5 &  6.0 & 22.8 \\
P2Net\cite{Wang2021P2NetJD}               & 55.2 & 46.7 & 13.0 & 36.2 & 32.0 & 32.8 &  5.8 & 31.7 \\
Predator\cite{Huang2020PREDATORRO}        & 34.7 & 33.8 & 16.6 & 25.9 & 23.1 & 22.2 &  7.5 & 23.4 \\
MATR\cite{Li20232D3DMATR2M}               & 72.1 & 66.0 & 31.3 & 60.7 & 50.2 & 52.5 & 18.1 & 50.1 \\
Diff$^2$I2P\cite{mu2025diff2i2p}          & 74.1 & 68.8 & 39.2 & 65.6 & 52.1 & 54.2 & 18.1 & 53.2 \\
\textbf{DMM-Align}                        & \textbf{76.2} & \textbf{69.3} & \textbf{42.9} & \textbf{66.7} & \textbf{52.1} & \textbf{55.6} & \textbf{18.9} & \textbf{54.5} \\
\midrule
\multicolumn{9}{c}{Feature Matching Recall $\uparrow$} \\
\midrule
FCGF\cite{Choy2019FullyCG}                &  99.7 &  98.2 &  69.9 &  97.1 &  83.0 &  87.7 & 16.2 & 78.8 \\
P2Net\cite{Wang2021P2NetJD}               & 100.0 &  99.3 &  58.9 &  99.1 &  87.2 &  92.2 & 16.2 & 79.0 \\
Predator\cite{Huang2020PREDATORRO}        &  91.3 &  95.1 &  76.7 &  88.6 &  79.2 &  80.6 & 31.1 & 77.5 \\
MATR\cite{Li20232D3DMATR2M}               & 100.0 &  99.6 &  98.6 & 100.0 &  92.4 &  95.9 & 58.1 & 92.1 \\
Diff$^2$I2P\cite{mu2025diff2i2p}          & 100.0 & \textbf{100.0} & 100.0 & 100.0 & 93.4 & 96.2 & 55.4 & 92.2 \\
\textbf{DMM-Align}                        & \textbf{100.0} & \textbf{100.0} & \textbf{99.9} & \textbf{100.0} & \textbf{93.7} & \textbf{96.9} & \textbf{57.1} & \textbf{92.5} \\
\midrule
\multicolumn{9}{c}{Registration Recall $\uparrow$} \\
\midrule
FCGF\cite{Choy2019FullyCG}                & 89.5 & 79.7 & 19.2 & 85.9 & 69.4 & 79.0 &  6.8 & 61.4 \\
P2Net\cite{Wang2021P2NetJD}               & 96.9 & 86.5 & 20.5 & 91.7 & 75.3 & 85.2 &  4.1 & 65.7 \\
Predator\cite{Huang2020PREDATORRO}        & 69.6 & 60.7 & 17.8 & 62.9 & 56.2 & 62.6 &  9.5 & 48.5 \\
MATR\cite{Li20232D3DMATR2M}               & 96.9 & 90.7 & 52.1 & 95.5 & 80.9 & 86.1 & 28.4 & 75.8 \\
Diff$^2$I2P\cite{mu2025diff2i2p}          & \textbf{99.0} & 95.6 & 74.0 & 98.9 & 86.8 & 90.2 & 36.5 & 83.0 \\
\textbf{DMM-Align}                        & 98.9 & \textbf{96.8} & \textbf{80.6} & \textbf{99.3} & \textbf{87.1} & \textbf{94.6} & \textbf{45.9} & \textbf{86.2} \\
\bottomrule
\end{tabular}
\end{table}

\begin{figure*}[t]
\vspace{-4pt}
  \centering
  \includegraphics[width=\textwidth]{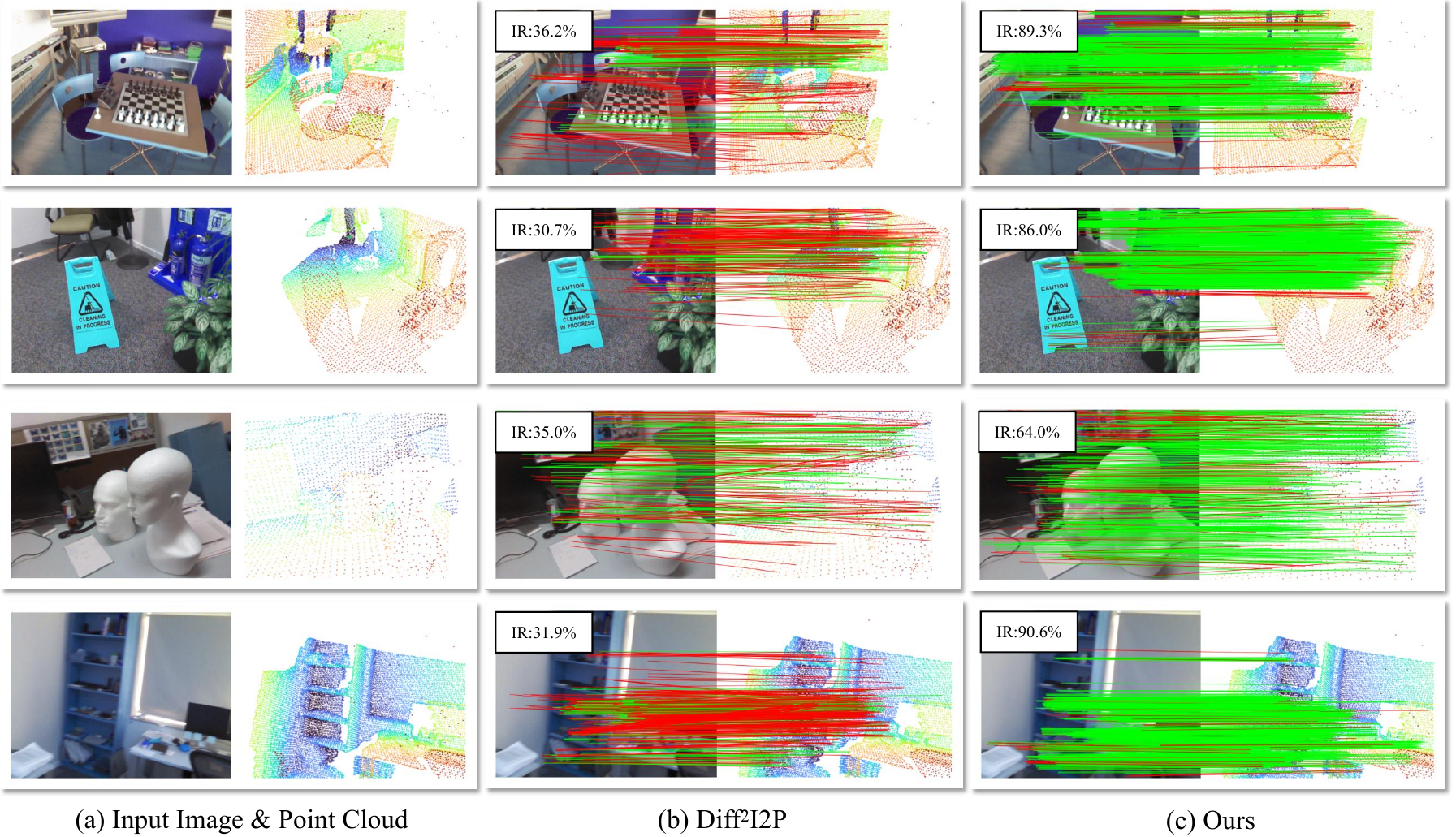}
  \caption{Qualitative results on the 7-Scenes dataset. The red lines indicate erroneous correspondences (3D distance greater than 5 cm), while the green lines represent correct correspondences.}
  \label{fig:3}
\vspace{-8pt}
\end{figure*}

\noindent\textbf{Qualitative results.}
Fig.~\ref{fig:3} visualizes representative correspondences produced by Diff$^2$I2P\cite{mu2025diff2i2p}  and DMM-Align on challenging examples from the 7-Scenes dataset. Column (a) shows the input image and point cloud pairs, while columns (b) and (c) present the results of Diff$^2$I2P\cite{mu2025diff2i2p}  and DMM-Align, respectively. Across all four examples, DMM-Align consistently yields cleaner and more spatially coherent correspondences, with substantially fewer outliers and higher inlier ratios. In the first two rows, which contain clutter, local ambiguity, and partial occlusion, Diff$^2$I2P\cite{mu2025diff2i2p}  produces many erroneous matches, resulting in relatively low inlier ratios of 36.2\% and 30.7\%. In contrast, DMM-Align suppresses mismatches effectively and improves the inlier ratios to 89.3\% and 86.0\%. Similar behavior can be observed in the last two rows, where repetitive structures and weak local distinctiveness make correspondence estimation particularly difficult. While the baseline suffers from heavy outlier contamination with inlier ratios of 35.0\% and 31.9\%, our method maintains much more reliable and geometrically consistent matches, improving the inlier ratios to 64.0\% and 90.6\%. These qualitative results support our quantitative findings and show that the proposed closed-loop design leads to more accurate and robust 2D--3D correspondence estimation in challenging scenes.

\subsection{Evaluations on RGB-D Scenes V2}
\noindent\textbf{Quantitative results.}
We further evaluate DMM-Align on RGB-D Scenes V2, which is more challenging due to larger viewpoint variation, heavier occlusion, and stronger cross-modal ambiguity. The results are reported in Tab.~\ref{tab:2}. DMM-Align achieves the best performance across all reported metrics, reaching 62.4\% in PIR, 39.8\% in IR, 78.4\% in FMR, and 63.7\% in RR. Compared with Diff$^2$I2P\cite{mu2025diff2i2p}, our method yields consistent improvements across all metrics, including gains of 1.6\% in PIR, 2.9\% in IR, 1.3\% in FMR, and 3.2\% in RR. The improvement over 2D3D-MATR\cite{mu2025diff2i2p} is even more pronounced, especially in PIR and RR. These results indicate that DMM-Align improves both coarse and fine correspondence estimation, leading to more stable matching and more robust pose estimation in challenging cross-scene settings. Together with the results on 7-Scenes, they further support our claim that the performance gain comes from the proposed closed-loop geometric feedback rather than dataset-specific tuning.

\begin{table}[h]
\caption{Registration results of DMM-Align and baselines on RGB-D Scenes V2. The best result for each metric is shown in \textbf{bold}.}
\label{tab:2}
\centering
\footnotesize
\setlength{\tabcolsep}{11.5pt}
\renewcommand{\arraystretch}{1.1}
\begin{tabular}{l|cccc}
\toprule
Method & PIR (\%)$\uparrow$ & IR (\%)$\uparrow$ & FMR (\%)$\uparrow$ & RR (\%)$\uparrow$ \\
\midrule
FCGF\cite{Choy2019FullyCG}               & 20.1 & 10.3 & 29.2 & 32.5 \\
P2-Net\cite{Wang2021P2NetJD}             & 30.4 & 14.5 & 63.7 & 41.7 \\
Predator\cite{Huang2020PREDATORRO}       & 32.6 & 15.8 & 68.1 & 33.6 \\
MATR\cite{Li20232D3DMATR2M}              & 57.6 & 36.3 & 76.0 & 56.9 \\
Diff$^2$I2P\cite{mu2025diff2i2p}         & 60.8 & 36.9 & 77.1 & 60.5 \\
\textbf{DMM-Align}                       & \textbf{62.4} & \textbf{39.8} & \textbf{78.4} & \textbf{63.7} \\
\bottomrule
\end{tabular}
\end{table}

\subsection{Ablation Studies}
We conduct ablation studies to analyze whether the gains of DMM-Align truly come from the proposed closed-loop design and to examine the contribution of each key component. Unless otherwise specified, all ablations are performed on 7-Scenes.

\noindent\textbf{Overall effect of DMM-Align.}
We evaluate the overall impact of DMM-Align by comparing the full model with several reduced variants in Tab.~\ref{tab:closedloop_main}. As shown in the table, DMM-Align consistently achieves the best performance across all evaluation metrics, reaching 54.5\% IR, 92.5\% FMR, and 86.2\% RR.
Compared with matrix diffusion only, the full model improves RR from 81.6\% to 86.2\%.
It also improves over the distillation-only variant from 83.3\% to 86.2\%, and over the uncoupled coexistence variant from 84.7\% to 86.2\%. These results show that DMM-Align provides a clear overall performance gain. Importantly, the full model performs better than both single-module variants and their uncoupled combination. This suggests that the improvement does not arise simply from introducing stronger individual components, but from explicitly coupling correspondence refinement, pose estimation, and representation learning through a differentiable geometric feedback pathway.

\begin{table}[h]
\caption{Overall effect of DMM-Align on 7-Scenes. DMM-Align consistently outperforms reduced variants, showing the benefit of explicit closed-loop coupling.}
\centering
\footnotesize
\setlength{\tabcolsep}{12pt}
\renewcommand{\arraystretch}{1.15}
\begin{tabular}{lccc}
\toprule
Model & IR $\uparrow$ & FMR $\uparrow$ & RR $\uparrow$ \\
\midrule
Matrix diffusion only & 52.7 & 92.1 & 81.6 \\
Distillation only & 53.3 & 92.3 & 83.3 \\
Coexistence without coupling & 53.9 & 92.4 & 84.7 \\
DMM-Align (full) & \textbf{54.5} & \textbf{92.5} & \textbf{86.2} \\
\bottomrule
\end{tabular}
\label{tab:closedloop_main}
\vspace{-10pt}
\end{table}

\noindent\textbf{Matching matrix modeling.}
We further analyze the effect of the matching matrix formulation in Tab.~\ref{tab:matrix_main}. A strict doubly stochastic assignment is too restrictive for practical 2D--3D registration, since real image--point-cloud pairs often contain occlusion, partial overlap, and unmatched elements. Relaxing the formulation to a partial doubly stochastic matrix improves performance by allowing more flexible cross-modal correspondence, and introducing a dustbin brings additional gains by explicitly modeling non-correspondence. On top of this formulation, geometry-aware diffusion yields the best results, improving IR from 53.8\% to 54.5\% and RR from 84.7\% to 86.2\%. These results suggest that the improvement does not come merely from a looser assignment or generic denoising, but from treating correspondence as a structured variable and refining it under geometric guidance, which leads to more robust matching and better final registration.

\begin{table}[h]
\centering
\caption{Effect of matching matrix modeling on 7-Scenes. More flexible correspondence modeling and geometry-aware diffusion consistently improve performance.}
\label{tab:matrix_main}
\footnotesize
\setlength{\tabcolsep}{6pt}
\renewcommand{\arraystretch}{1.15}
\begin{tabular}{lccc}
\toprule
Matching formulation & IR $\uparrow$ & FMR $\uparrow$ & RR $\uparrow$ \\
\midrule
Strict doubly stochastic & 52.6 & 92.0 & 82.4 \\
Partial doubly stochastic & 53.1 & 92.1 & 83.6 \\
Partial DS + dustbin & 53.8 & 92.3 & 84.7 \\
Partial DS + dustbin + geometry-aware diffusion & \textbf{54.5} & \textbf{92.5} & \textbf{86.2} \\
\bottomrule
\end{tabular}
\end{table}

\begin{figure*}[t]
\vspace{-4pt}
  \centering
  \includegraphics[width=\textwidth]{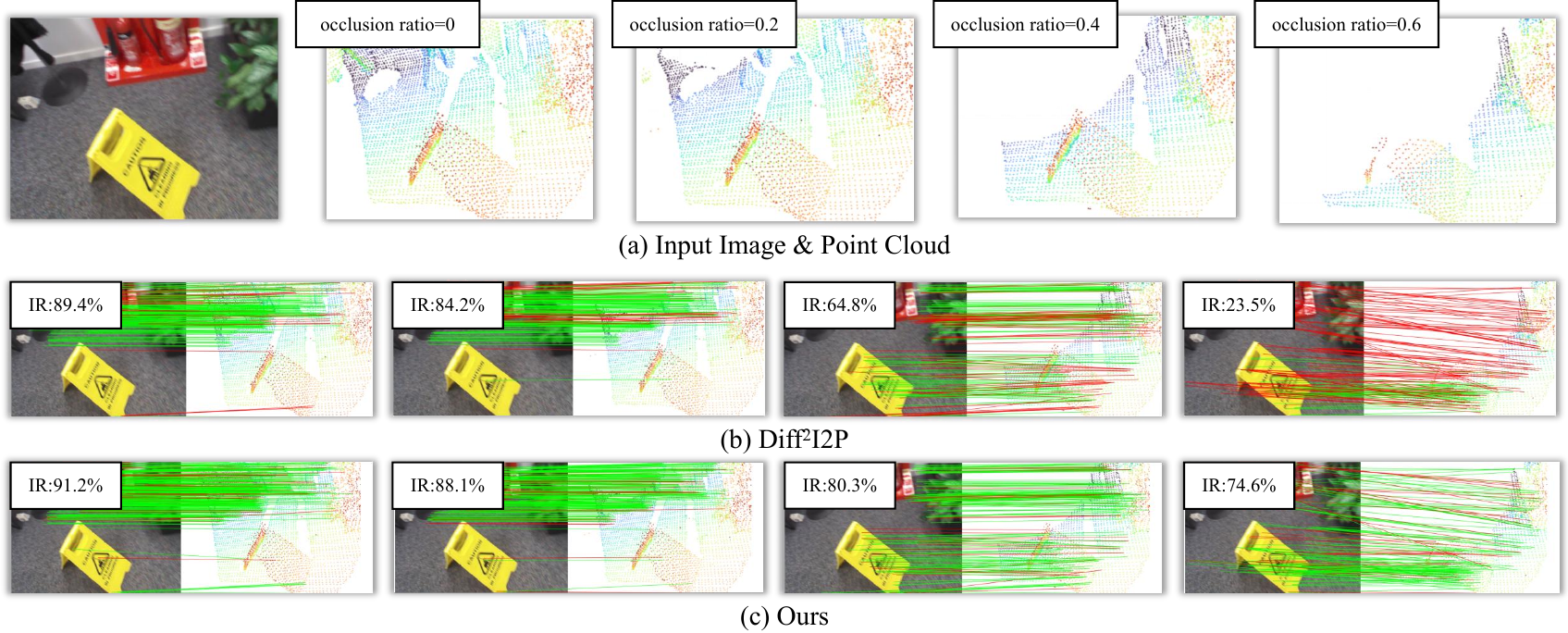}
  \caption{
\textbf{Robustness under local point cloud degradation.}
From left to right, the occlusion ratio increases from $0$ to $0.6$. (a) Input image and degraded point cloud. (b) Diff$^2$I2P. (c) DMM-Align. Green and red lines denote correct and incorrect matches, respectively, and IR is reported for each case. DMM-Align consistently achieves higher IR under severe degradation.}
\label{fig:4}
\vspace{-8pt}
\end{figure*}

\noindent\textbf{Differentiable geometric hinge.}
We further analyze the effect of the differentiable geometric hinge in Tab.~\ref{tab:hinge_main}. Without the hinge, the connection between correspondence refinement and pose estimation becomes weaker, leading to lower matching and registration performance. Introducing 3D residual correction improves the results by allowing local geometric adjustment before pose estimation. Adding 2D correction further enhances alignment quality, and uncertainty-aware weighting yields the best overall performance, improving IR from 54.1\% to 54.5\% and RR from 85.5\% to 86.2\%. These results suggest that the hinge is not merely an auxiliary design, but a key component that lifts local correspondence updates into a trainable global pose state and makes the closed-loop optimization more effective.

\begin{table}[h]
\centering
\caption{Effect of the differentiable geometric hinge on 7-Scenes. Residual correction and uncertainty-aware weighting consistently improve performance.}
\label{tab:hinge_main}
\footnotesize
\setlength{\tabcolsep}{6pt}
\renewcommand{\arraystretch}{1.15}
\begin{tabular}{lccc}
\toprule
Hinge variant & IR $\uparrow$ & FMR $\uparrow$ & RR $\uparrow$ \\
\midrule
Without hinge & 53.2 & 92.1 & 83.8 \\
+ 3D offset only & 53.7 & 92.2 & 84.7 \\
+ 3D offset + 2D offset & 54.1 & 92.4 & 85.5 \\
+ 3D offset + 2D offset + uncertainty weighting & \textbf{54.5} & \textbf{92.5} & \textbf{86.2} \\
\bottomrule
\end{tabular}
\vspace{-8pt}
\end{table}

\noindent\textbf{Geometry-conditioned distillation.}
We further analyze the effect of geometry-conditioned distillation in Tab.~\ref{tab:distill_main}. Using only depth with a binary mask gives the weakest result, indicating that coarse geometric supervision alone is insufficient for reliable feature learning. Replacing the binary mask with soft masking and adding reprojection cues improves performance, and using the full geometry bundle further strengthens both matching and registration quality. The best results are achieved with uncertainty-aware soft masking, which improves IR from 54.0\% to 54.5\% and RR from 85.2\% to 86.2\%. These results suggest that the benefit of the distillation module does not come merely from adding a stronger supervisory branch, but from injecting structured and confidence-aware geometric feedback into the representation space, thereby making the closed-loop optimization more effective.

\begin{table}[h]
\centering
\caption{Effect of geometry-conditioned distillation on 7-Scenes. Richer geometric cues and uncertainty-aware masking consistently improve performance.}
\label{tab:distill_main}
\footnotesize
\setlength{\tabcolsep}{4pt}
\renewcommand{\arraystretch}{1.15}
\begin{tabular}{lccc}
\toprule
Distillation variant & IR $\uparrow$ & FMR $\uparrow$ & RR $\uparrow$ \\
\midrule
Depth only + binary mask & 52.5 & 92.0 & 82.4 \\
Depth + reprojection + soft mask & 53.3 & 92.2 & 83.9 \\
Full geometry bundle + soft mask & 54.0 & 92.4 & 85.2 \\
Full geometry bundle + uncertainty-aware soft mask & \textbf{54.5} & \textbf{92.5} & \textbf{86.2} \\
\bottomrule
\end{tabular}
\end{table}

\section{Conclusion}
We presented DMM-Align, a closed-loop framework for detection-free 2D--3D registration. By coupling correspondence refinement, pose estimation, and representation learning through differentiable geometric feedback, DMM-Align improves both matching quality and final registration accuracy. Experiments on 7-Scenes and RGB-D Scenes V2 show consistent gains in accuracy and robustness, especially under low overlap, occlusion, and strong ambiguity. These results suggest that robust 2D--3D registration benefits not only from stronger modules, but also from a better optimization structure.

\begin{acks}
This work was supported by the National Natural Science Foundation of China (62373164), the Natural Science Foundation of Shandong Province (ZR2025QC2246Z), the Taishan Scholar Foundation of Shandong Province (tsqn202507271), the Central Government Guides Local Program (YDZX2024075), and the Taishan Experts Program (tscy20241154).
\end{acks}



\bibliographystyle{ACM-Reference-Format}
\bibliography{main/sample-base}










\end{document}